\documentclass[fullpaper,cameraready]{nldl}
\renewcommand{\DOI}{10.7557/18.6301}
\usepackage[utf8]{inputenc}
\usepackage{url}
\usepackage{graphicx}
\usepackage{authblk}
\usepackage{amsmath}
\usepackage{algorithm}
\usepackage{algorithmic}
\usepackage{hyperref}
\usepackage{cleveref}

\usepackage{xspace,amsfonts,amsthm,cleveref,thm-restate}
\usepackage{todonotes}

\newtheorem{definition}{Definition}

\renewcommand{\t}{t}
\newcommand{\HORN}{LRN\xspace}
\newcommand{\HORNs}{\ensuremath{{\sf LRN}^\ast}\xspace}

\usepackage{color, colortbl}
\definecolor{Gray}{gray}{0.93}

\newcommand{\MQc}[2]{\ensuremath{{\sf MQ}_{#1,#2}\xspace}}
\newcommand{\EQc}[2]{\ensuremath{{\sf EQ}_{#1,#2}\xspace}}

\newcommand{\NN}{\ensuremath{N}\xspace}
\newcommand{\encoding}{\ensuremath{{\sf vector}}\xspace}
\newcommand{\targetNN}{\ensuremath{t_\NN}\xspace}

\newcommand{\target}{\ensuremath{{ t}}\xspace}
\newcommand{\hypo}{\ensuremath{{h}}\xspace}

\newcommand{\formula}{\ensuremath{\phi}\xspace}

\newcommand{\ant}{\ensuremath{{\sf ant}}\xspace}
\newcommand{\con}{\ensuremath{{\sf con}}\xspace}

\newcommand{\mn}[1]{\ensuremath{\mathsf{#1}}}

\newlength{\indxlength}
\newcommand{\Amc}{\ensuremath{\mathcal{A}}\xspace}

\newcommand{\Dmc}{\ensuremath{\mathcal{D}}\xspace}
\newcommand{\Emc}{\ensuremath{\mathcal{E}}\xspace}
\newcommand{\Fmc}{\ensuremath{\mathcal{F}}\xspace}

\newcommand{\Hmc}{\ensuremath{\mathcal{H}}\xspace}
\newcommand{\Imc}{\ensuremath{\mathcal{I}}\xspace}

\newcommand{\Kmc}{\ensuremath{\mathcal{K}}\xspace}
\newcommand{\Lmc}{\ensuremath{\mathcal{L}}\xspace}

\newcommand{\Nmc}{\ensuremath{\mathcal{N}}\xspace}

\newcommand{\Rmc}{\ensuremath{\mathcal{R}}\xspace}

\newcommand{\Fmf}{\ensuremath{\mathfrak{F}}\xspace}

\newcommand{\Rsf}{\ensuremath{\mathsf{R}}\xspace}

\newcommand{\Vsf}{\ensuremath{\mathsf{V}}\xspace}

\newcommand{\csf}{\ensuremath{\mathsf{c}}\xspace}

\newcommand{\fsf}{\ensuremath{\mathsf{f}}\xspace}

\newcommand{\psf}{\ensuremath{\mathsf{p}}\xspace}

\newcommand{\rsf}{\ensuremath{\mathsf{r}}\xspace}

\newcommand{\usf}{\ensuremath{\mathsf{u}}\xspace}
\newcommand{\vsf}{\ensuremath{\mathsf{v}}\xspace}

\include{misc}
\usepackage[square,numbers]{natbib}
\title{Extracting Rules from Neural Networks  \\ with Partial Interpretations}
\author{Cosimo Persia, Ana Ozaki}
\affil{Department of Informatics\\ University of Bergen, Norway}
\date{\vspace{-5ex}}

\begin{document}
\nldlmaketitle

\begin{abstract}  
We investigate the problem of  extracting rules,
expressed in Horn logic, from neural network
models. % by posing queries. 
Our work is based on the exact learning model, in which a learner interacts with a
teacher (the neural network model) via %membership and equivalence 
queries in order to learn an abstract
target concept, which in our case is a set of Horn rules. %A membership query
We consider \emph{partial interpretations} to formulate
%membership queries and counterexamples
the queries. These can be understood as a representation
of the world where part of the knowledge regarding the truthness of
 propositions is unknown. 
We employ Angluin’s algorithm
for learning Horn rules via queries %based on partial interpretations
 and evaluate our strategy empirically. 
%Then, we extend our approach
%to the case in which rules are expressed in possi-
%bilistic logic—a formalism designed for dealing
%with uncertainty—and evaluate it empirically.
\end{abstract}

\section{Introduction}

Neural networks have been used to achieve important 
milestones in artificial intelligence~\cite{DBLP:journals/ai/CampbellHH02,alphago,DBLP:conf/icml/LeRMDCCDN12,watson},
but it is difficult to understand how predictions of the models are made,
 %based on the input example. 
and this limits their usability.
In this work, we propose an approach 
for extracting rules from
black-box machine learning models, such as neural networks.
It is often the case that not all values in a dataset 
are known or trustable.
For this reason, our approach assumes settings in which the dataset used to train the neural network 
contains missing values.

We first binarize a given dataset and we train a neural network with it.
Then, we run the \HORN algorithm~\cite{Frazier1993LearningFE}.
This algorithm poses queries  to the neural network,
seen as a teacher, in order to extract
  rules encoded in it.
Rules are represented using Horn logic, for example, they can be of the form
$ (({\sf horse} \land {\sf wings}) \rightarrow {\sf pegasus}) $.
With Horn rules, we can carry  automated reasoning
  in polynomial time, and it is feasible to check the quality of the model.
 
 We perform 
 an empirical study
 using the 
 hepatocellular carcinoma (HCC) dataset~\cite{SANTOS201549},
which describes survivability of patients 
diagnosed with hepatocellular carcinoma
according to clinical information.
HCC contains many missing values of attributes of patients.
 We compare the hypothesis built
with our approach with the hypothesis built by a state-of-the-art 
implementation of the incremental decision tree algorithm~\cite{DBLP:conf/kdd/DomingosH00}.
Our rule extraction procedure 
correctly extracts meaningful rules and it
is two times faster than the 
decision tree algorithm.

%\nb{todo: to improve this} 
%Interpretations assign a truth value to each proposition.
%It is often  the case that  the truth values of some propositions
%is not known. Partial interpretations can express this with a third symbol (in this work $\ast$)
%that indicates lack of knowledge.  
%Here, we consider the problem of extracting Horn rules from 
%neural networks trained with partial interpretations.

\noindent
{\bfseries Related Work.}
A similar work~\cite{DBLP:conf/icml/WeissGY18} extracts probabilistic  automata from neural networks by asking queries,
and a recent work~\cite{DBLP:conf/aaai/OkudonoWSH20} focuses on how to better simulate  queries asked to black-box models.
We can also find methods that verify binarized neural networks 
by extracting a binary decision diagram~\cite{10.1007/978-3-030-24258-9_25} through queries.
The interpretability field is large and there are many approaches to interpret neural networks models \cite{9521221}.
Our technique belongs to the global and active approach that explains the already trained model as a whole,
as opposed to changing the network architecture for interpretability (passive), or explaining through feature studies or correlation (local).

\section{Preliminaries}

We provide basic definitions for (propositional) logic
and the exact learning model. 
%in particular Horn logic, and the
%learning models used.
% , and the algorithm
%for learning propositional Horn theories. % and learning theory.

\subsection{Logic and Neural Networks}

Let $ \Vsf$ % = \{ \vsf_1,\cdots,\vsf_n \} $ 
be a finite set of \emph{boolean variables}.
\emph{A literal} over \Vsf is either a \emph{variable} $ \vsf \in \Vsf $ or its negation $ \neg \vsf $.
A literal   is  \emph{positive}, if it is a variable, 
and \emph{negative} otherwise. 
%negaative,$ \neg $ does not appear in \lsf,
%otherwise it is said to be \emph{negative}. 
\emph{A  clause}  over \Vsf  is a disjunction ($\vee$) of literals over \Vsf.
%\nb{new here}
It is  \emph{Horn} 
%clause %or \emph{Horn formula} 
%is a clause 
%where 
if at most one literal is positive.
%i.e. of the form
%$ \lsf_1 \lor \cdots \lsf_m $  $ \lsf_i $ where for each $ 1\leq i \leq m $.
%$ \lsf_i $ is a literal over \Vsf
%where 
% The part of the clause that contains only negative literals is called  and the part with at most one positive literal is called \emph{head}.
%~ \nb{not sure if the ex is needed} % Maybe we could avoid talking about material implication and just rely on the semantics
%~ \begin{example}\label{ex:clauses} 
%~ Let $ \csf = \neg \vsf_1 \lor \neg \vsf_2  \lor \vsf_3$ be a horn clause.
%~ By the rule of `material implication',
%~ the formulas $ \phi \rightarrow \psi $ and
%~ $ \neg \phi \lor \psi $
%~ are equivalent.
%~ It follows that we can write \csf 
%~ as 
%~ $ (\vsf_1\land \vsf_2 )\rightarrow \vsf_3 $.
%~ %	
%~ Moreover,
%~ $ \neg \vsf_1 \lor  \vsf_2  \lor \vsf_3$ is not a Horn clause because it has more than one positive  literal.
%~ The Horn clause $ \neg \vsf_1 \lor \neg \vsf_2  \lor \neg \vsf_3$ 
%~ corresponds to the rule $(\vsf_1\land \vsf_2 \land\vsf_3)\rightarrow \bot$. %is a horn clause.
%~ \hfill {\mbox{$\triangleleft$}} 
%~ \end{example}
\emph{A (propositional) formula} over \Vsf is a conjunction of clauses over \Vsf (in conjunctive normal form). 
It is  \emph{Horn} if its clauses are Horn. %and Horn theories as conjunction of implications

%

%An interpretation \Imc  over \Vsf is a function that maps every variable in \Vsf
%to either $0$ (false) or $1$ (true).
%We denote by $ \Imc_\top $ the subset of variables mapped to true. \nb{new}
An \emph{interpretation} is a function %partial interpretation
that maps  %such subset is \Vsf, that is, 
all variables \Vsf %are mapped
to either $0$ (false) or $1$ (true). %$\{0,1\}$. 
It also  maps the constant symbol $ \top $ %(always true) 
to $1$
and $ \bot $ %(always false) 
to $0$.
We   write $v\in\Imc$ if
$\Imc(v)=1$. %\nb{this was already there} ok great!
We may omit `over \Vsf' in formulas, clauses,  literals, and interpretations. %\nb{theories -> formulas}.
A variable $ \vsf\in\Vsf $ is \emph{satisfied} by \Imc 
if $ \vsf \in \Imc $, otherwise it is \emph{falsified}.
A negative literal $ \neg \vsf $ is satisfied by \Imc iff 
\vsf is falsified by \Imc.
A   clause \csf is satisfied by \Imc iff at least one literal in $ \csf $ is satisfied by \Imc
and a   formula  $ \t $ is satisfied by \Imc iff  
every clause in $ \t $ is satisfied by \Imc.
%\nb{P: updated until the end of this paragraph}
A \emph{partial interpretation}  extends the notion of an interpretation by
allowing some values to be ``missing'' or ``unknown'', denoted `$?$'. In detail,
 it is a function  which maps
%a \emph{subset} of 
\Vsf  to $\{0,1,?\}$. %, with  $?$ representing
A partial interpretation \Imc satisfies a formula $ \t $
if there is a way to replace each $?$ in its image by either 
$0$ or $1$ and the resulting function satisfies $ \t $. 
%
%either $0$ (false) or $1$ (true).
 
%We  treat a set of variables and the conjunction of its elements interchangeably.
Horn clauses $\csf$ can be written as \emph{rules} %implications 
of the form   %\nb{P: new sentence}
$ \ant(\csf)\rightarrow \con(\csf) $, where $ \ant(\csf) $ (\emph{antecedent}) 
is the set of variables that occur negated in \csf 
or the constant symbol $ \top $   if none is negated; and
$ \con(\csf) $ (\emph{consequent}) is the positive literal  in \csf
or $ \bot $   if none is positive.

% (\Cref{ex:interpretation}). 

%I think it is not needed
%~ \begin{example}\label{ex:interpretation}
%~ Let $ \csf_1 = (\vsf_1\land \vsf_2) \rightarrow \vsf_3 $, \
%~ $ \csf_2 = (\vsf_3\land \vsf_2) \rightarrow \vsf_1 $, and
%~ $ \csf_3 = \vsf_1\rightarrow \vsf_2 $.
%~ The interpretation $ \Imc = \{ \vsf_1, \vsf_2 \} $
%~ falsifies $ \csf_1 $ and it 
%~ satisfies $ \csf_2 $ and $ \csf_3 $.
%~ \hfill {\mbox{$\triangleleft$}} 
%~ \end{example}

If an interpretation \Imc satisfies a literal, a   clause or theory $ x $,
we write $ \Imc \models x $, otherwise, $ \Imc \not\models x $.
Let $ \t  $ be a   theory and let $ \csf $ be a   clause.
If, for every  \Imc, we have that $ \Imc \models \t $
implies $ \Imc \models \csf $,
then we write $ \t \models \csf $ %(meaning that $ \csf $ is a logical consequence of \Kmc) 
and we say that $ \t $  \emph{entails} \csf.
If $ \t  $ entails every clause in a   theory $ \t' $,
then we also write $ \t \models \t' $.
If $ \t'\models \t $ also holds,
then $ \t $ and $ \t' $ are logically equivalent and we write
$ \t \equiv \t' $.
We say that a formula $\formula$ is \emph{satisfiable} if there is 
an interpretation $\Imc$ such that $\Imc \models \formula$
%.
%Moreover, $\formula$ is 
and \emph{falsifiable} if its negation $\lnot \formula$ is satisfiable.
%An Horn theory 
%Moreover, $\formula$ is \emph{non-trivial} 
%if it is satisfiable and falsifiable.

%
%Our neural networks %are essentially binary acceptors 
%are trained on a set of classified
%Oracles in this work are simulated by  
Neural network models in this work can be understood as an alternative
way of representing  a formula in propositional logic.
%which are  binary classifiers. 
A neural network model \NN is a function
that receives a vector    in the  $ |\Vsf|  $ dimensional space,
with values in $\{0,1,?\}$ (with `?' standing for an `unknown value'), %\nb{Should we also consider the unknown value?}
and outputs a classification of this input.
The mapping from interpretations to vectors is defined as follows.
Given an interpretation \Imc over \Vsf,
we assume a total order on the elements of \Vsf and
%and its output is either $ 1 $ or $ 0 $.
denote by $ \encoding(\Imc) $ 
the vector in the  $ |\Vsf|  $ dimensional space
where the element %$\vsf_i$ 
at position $ i$ %\in\{1, \cdots,n \} $
is $ 1 $ if $ \vsf_i \in \Imc $ and $ 0 $ otherwise.
In this work, a \emph{dataset} is a set of elements of the form
$ (\encoding(\Imc),l) $,
where $l$ is either $0$ or $1$ and \Imc is a partial
 interpretation.
For every
neural network \NN trained on a given dataset, 
there is a propositional formula  
%it will have 
%encoded a   theory 
\targetNN
such that 
$ \NN(\encoding(\Imc)) = 1 $
iff 
$ \Imc\models \targetNN $.
In this sense, \NN 
can be seen as an alternative representation of
\targetNN.
%In Section, this classification can only be $ 1 $ or $ 0 $
%In Section~\ref, it is a value in the interval $[0,1]$.   

\subsection{Learning via Queries}

To formally define the problem setting, % learning problem under consideration, 
%we formally define the objects of interest when studying learning problems
we use the notion of a 
%\emph{learning framework} \Fmf over  a set of variables \Vsf.
\emph{learning framework} \Fmf 
%We define \Fmf 
as pair $ (\Emc,\Hmc) $, where 
%\Emc is the set of interpretations over \Vsf
%and \Hmc is the set of theories over \Vsf.
\Hmc is the set of all formulas
in propositional logic %\nb{it does not make sense to introduce prop logic and then talk about FO}
%with a
%first-order (FO) logic language \cite{BellJ.L.JohnLane1977Acim}
and \Emc is the set of all partial interpretations (over variables in \Vsf). %of that language.
We say that \Fmf is \emph{Horn} if \Hmc
is restricted to the set of all Horn formulas. %theories over 
%Also,   \Fmf is \safe if $l\in\Hmc$ implies that $l'\in\Hmc$, for all $l'\subseteq l$.
%(over \Vsf).
%a set of variables \Vsf that are Horn.
%and \Emc is the set of interpretations over \Vsf.
%Examples in \Emc characterise elements
%in \Hmc. 
%We say that $\Fmf$ 
%is \emph{non-trivial} 
%if $\Hmc$ contains a non-trivial FO KB; and 
%that it 
%It is \emph{\safe} if  $h\in\Hmc$ implies that
%$h'\in \Hmc $, for all $h'\subseteq h$.
%In particular, we have that a Horn learning framework is also \safe. 
For any $ h\in\Hmc $, 
%we say that 
$ \Imc \in\Emc $ is a \emph{positive example for $ h $}, 
if $ \Imc \models h $, and, a \emph{negative example for $ h $} if $ \Imc \not\models h $.
For any  $ h,t\in \Hmc $,
a \emph{counterexample}
for $ t $ and $ h $ is an example $ \Imc \in \Emc $
such that either $ \Imc \models t $ and $ \Imc \not\models h $ (a \emph{positive counterexample}), or $ \Imc \models h $ and $ \Imc \not\models t $ (a
\emph{negative counterexample}).

%To learn a theory from a teacher (represented as neural network), we consider the exact learning model~\cite{angluinqueries}. 
%More specifically, 
We study the problem of  identifying an unknown 
target $ t\in \Hmc $ by  posing queries to two kinds of oracles~\cite{Frazier1993LearningFE}
(implementation in Section~\ref{ss:lrn}).
A \emph{membership  oracle} \MQc{\Fmf}{t} is a function 
that takes as input $ \Imc\in\Emc $ and it outputs
`yes' if $ \Imc \models t $, `no' otherwise.
An \emph{equivalence  oracle} \EQc{\Fmf}{t} 
%is a function that 
takes as input a hypothesis $ h\in\Hmc $
and it outputs `yes' if $ h\equiv t $,
otherwise, it outputs a counterexample for $ t $ and $ h $.
%We consider two types of queries: \emph{membership query}
%and \emph{equivalence query}.
A \emph{membership query} is a call to \MQc{\Fmf}{t}
and an \emph{equivalence query} is a call to \EQc{\Fmf}{t}.
%\nb{introduce learning systems?}
%%\nb{this paragraph may be removed}
\begin{definition}[Exact Learning]\upshape \label{d:exactlearning}
	A learning framework \Fmf $ (\Emc,\Hmc) $ is \emph{exactly learnable} 
	if 
	%\begin{itemize}
	%\item 
	there is a deterministic algorithm $A$ that takes as input
	the set of variables \Vsf
	used to formulate the target $ t \in\Hmc$,
	asks membership and equivalence queries,
	and   outputs a hypothesis $ h \in\Hmc$ equivalent to $t$.
	%\end{itemize}
	We say that \Fmf is \emph{ exactly learnable in polynomial time} if
	the 
	number of steps used by $A$ is bounded by a polynomial
	on $ |t|$
	and the largest counterexample seen so far. Each query counts as one step of computation.
\end{definition}

\section{Extracting Horn Rules with Partial Interpretations}
The goal of our work is to find rules hidden in a 
black box machine learning model such as a
trained neural network model.
We %assume that the neural network expresses Horn rules and we 
present %to use
an adaptation of
the \HORN algorithm~\cite{Frazier1993LearningFE} that learns from 
partial interpretations instead of entailments, as originally proposed by the authors of the mentioned paper. %\nb{O: changed here, may need improvement}
This algorithm is able to exactly identify %in polynomial time 
any unknown target Horn theory
%By assuming the existence of 
by posing queries to oracles that can answer membership and equivalence queries. 
The algorithm is guaranteed to terminate in 
polynomial time 
with respect to the number of variables into consideration.

\subsection{The $ {\sf \HORN}^\ast $ Algorithm}
\label{ss:lrn}

We adapted \HORN so that it is able to learn rules from partial interpretations.
Membership queries take as input 
partial interpretations and counterexamples to equivalence queries  are also partial interpretations.
\Cref{a:horn} shows the main steps of the modified algorithm.
\begin{algorithm}[]
	\begin{algorithmic}[1]
		\STATE {\bfseries Input:} It is assumed that the learner knows \Fmf (that is, it knows
		that the hypothesis should be a Horn theory) but not the target $t$. %\Vsf: variables.
		\STATE {\bfseries Output:} $h$ such that $h\equiv t$.
		
		\STATE Let $ S $ be the empty sequence. 
		\STATE Denote with $ \Imc_i $   the $ i $-th element of $ S $. %}}
		\STATE Let $ \hypo $ be the empty hypothesis.
		\WHILE {$ \EQc{\Fmf}{\target} (h)$ returns a counterexample $ \Imc $}
		%			\STATE \COMMENT{positive counterexample}
		%		\IF{$ \Imc\not\models \hypo $}
%		\IF{there is $ \csf\in h $ such that $ \Imc\not\models \csf $}
%		\STATE 	remove all $ \csf\in \hypo $ such that  $ \Imc\not\models \csf $ 
%		\ELSE
		%			\STATE\COMMENT{negative counterexample}
		%		\ELSE 
		\IF{there is $\Imc_i\in S$ such that $\Imc_i \cap \Imc \subset \Imc_i$ and  
			$\phantom{aa}$ $\MQc{\Fmf}{\target}(\Imc_i \cap \Imc) =   \text{`no'}$
		}
		%		\STATE %Find the first 
		%		$ i = \min\limits_{\Imc_j\in S} \{ j \mid   \Imc_j \cap \Imc \subset \Imc_j, 
		%		 \MQc{\Fmf}{\target}(\Imc_j \cap \Imc) =   \text{`no'}$
		\STATE replace the first such $ \Imc_i $ with $ \Imc_i \cap \Imc $ in $ S $
		%		\FOR{$ \Imc_i\in S $ s.t. $ \Imc_i \cap \Imc \subset \Imc_i $}
		%%		\STATE $ \gamma :=  \PQc{\Fmf_\pi}{\target}((\Imc_i \cap \Imc))   $
		%		%					\STATE $ p = \p(\alpha) $
		%		\IF{$ \MQc{\Fmf_\pi}{\target}(\Imc_i \cap \Imc) =  $ `no'}
		%		\STATE replace the first $ \Imc_i $ with $ \Imc_i \cap \Imc $ in $ S $
		%		\STATE {\bfseries break}
		%		\ENDIF
		%		\ENDFOR
		\ELSE
		%		\IF{no element in $ S $ has been replaced}
		\STATE append $ \Imc $ to $ S $
		\ENDIF
		
		%		\STATE $ \hypo := \emptyset $
		
		%		\FOR{$ \Imc\in S $}
		\STATE   $\hypo := \bigcup\limits_{ \Imc\in S }   \{
		%\bigwedge_{\usf\in(\Vsf\cup\{\bot\})\setminus\Imc}
		%		(
		(\bigwedge\limits_{\vsf\in \Imc} \vsf )\rightarrow \usf \mid \usf\in {\sf RHS}(\Vsf, \bigwedge\limits_{\vsf\in \Imc} \vsf)\}$
		
		%	\STATE   $\hypo := \bigwedge\limits_{ \Imc\in S }  \bigwedge_{\usf\in(\Vsf\cup\{\bot\})\setminus\Imc}
		%		((\bigwedge_{\vsf\in \Imc\cup\{\top\}} \vsf )\rightarrow \usf )$
		%		to \hypo
		%		\ENDFOR
		
%		\ENDIF
		\ENDWHILE
		
		\STATE {\bfseries return} \hypo
	\end{algorithmic}
	\caption{$ {\sf LRN}^\ast $}
	\label{a:horn}
\end{algorithm}

\begin{algorithm}[]
	\begin{algorithmic}[1]
		\STATE {\bfseries Input:} \Vsf: variables. $ \alpha \subseteq \Vsf $ 
		\STATE {\bfseries Output:} A subset of $ \Vsf \cup \{\bot \} $
		
		\STATE {\bfseries return} $ \{ v \mid v \in \Vsf\cup\{\bot\}\setminus\alpha,  \  \MQc{\Fmf}{\target}(\bigwedge\limits_{u\in  \alpha}
		u \rightarrow v) =   \text{`yes'}  \} $
	\end{algorithmic}
	\caption{$ {\sf RHS} $}
	\label{a:horn}
\end{algorithm}

\HORN 
poses  equivalence queries until it receives `yes' as an  answer. 
It keeps track of important partial interpretations that falsify the target.
Each such partial interpretation corresponds to  a rule entailed by the target~\cite{DERAEDT1997187}.
Upon receiving a negative counterexample,
the algorithm asks membership queries to 
find more specific antecedents of rules entailed by the target.
After that, it adds to the hypothesis rules entailed by the target by asking membership queries 
and the process repeats.
Correctness and termination of \Cref{a:horn} 
can be proven simi`larly as
with the \HORN algorithm~\cite{Frazier1993LearningFE}.
This is possible because we
can simulate membership and equivalence queries from the learning from entailments setting 
to the learning from partial interpretations setting (and vice-versa)~\cite[Theorem~16]{DERAEDT1997187}. 
%(learning from partial interpretations).

%
%We would like to \emph{learn rules from the neural network model}
%by posing queries. 
%
To simulate the membership oracle \MQc{\Fmf}{\targetNN}, we directly use the classifier \NN.
Whenever the algorithm calls \MQc{\Fmf}{\targetNN} with input a partial interpretation \Imc,
we check if %make \NN classify \Imc. If 
$ \NN(\encoding(\Imc)) = 1 $, which  means that 
$ \Imc\models \targetNN $. If so, we return the answer `yes' to the algorithm,
`no' otherwise.
Simulating an equivalence query oracle \EQc{\Fmf}{\targetNN} is not as straightforward
as we are checking if the hypothesis constructed is equivalent to \targetNN.

We   simulate \EQc{\Fmf}{\targetNN} 
by generating a set of examples randomly and classifying the examples using membership queries.
Then, we can search for examples in this set that the 
hypothesis constructed by 
\HORN 
misclassifies.
Depending on the size of the  set of examples randomly generated~\cite[Section~2.4]{angluinqueries},
if the hypothesis does not misclassify any example then
one can ensure  that with high probability the total number 
%there 
%will not be many 
interpretations misclassified (considering the entire space of partial interpretations) is low.
% between 
%\targetNN and the hypothesis.
More precisely, if the size of the set of examples generated randomly is at least
$ \frac{1}{\epsilon} log_2(\frac{{|\Hmc|}}{\delta})\text{~\cite{vapnik:264}}, $
 then one
can ensure that 
the hypothesis constructed is {probably approximately correct}~\cite{Valiant}.
The parameter $ \epsilon \in (0,1)$ 
indicates the probability that
the hypothesis misclassifies an interpretation
w.r.t. the target
and $ \delta \in (0,1)$ is
the probability that the learned hypothesis
errs more than $ \epsilon $. 

If $\Hmc$ 
corresponds to the class of formulas 
only expressible with Horn logic and variables 
 \Vsf, then 
 the number of
 logically different hypothesis in \Hmc is
 close to~\cite{alekseev,sztaki427}:
\begin{equation}\label{sizesample}
2^{\binom{|\Vsf|}{\left\lfloor|\Vsf|/2\right\rfloor} }. 
\end{equation}
This number follows from the fact that 
Horn logic is closed under intersection: if $\Imc$ and $\Imc'$
satisfy a Horn theory then $\Imc\cap\Imc'$ also does \cite{horn}.

\subsection{Representing constraints}

We explain how we can express
constraints that are going to be extracted in the experimental section.
Horn rules $ r $ are of the form $ \ant(r) \rightarrow \con(r) $:
$ ({\sf sunny} \land {\sf happy }) \rightarrow { \sf jogging}  $
where all the variables both in the antecedent and in the consequent are not negated.
This means that 
with 
Horn logic 
we cannot express rules of the form:
\begin{equation*}
\begin{array}{l}
	(\neg {\sf sunny} \land {\sf happy }) \rightarrow {\sf boardgame\_night}. \\
	({\sf empty\_fridge} \land {\sf hungry }) \rightarrow \neg {\sf happy }.
\end{array}
\end{equation*}

To express a `weak' form of negation, we   duplicate all the variables in \Vsf and treat every new variable as  the negation of a variable in \Vsf.
For example, let
$ \hat{\vsf}_i $ be the duplicated variable of  any $ \vsf_i\in\Vsf $.
We can express the rule 
\[ (\hat{{\sf sunny}}_1 \land {\sf happy }) \rightarrow {\sf boardgame\_night}. \]
%(that means "not $ {\sf sunny} $ and $ {\sf happy} $ imply $ {\sf boardgame\_night} $").
Usually, when duplicating variables in this way,
we would like to avoid that both paired variables are true in a partial interpretation
%to %always have disjoint values 
(since they represent each other's  negation).
For this reason, we assume that Horn rules of the form %\nb{added HORN}
\begin{equation}\label{dualconstraint}
	( \vsf \land \hat{\vsf} ) \rightarrow \bot
\end{equation}
always hold, for every $ \vsf\in\Vsf $.

\section{Experiments}

In this section we show experimental results  using the  approach presented in the previous section
where  a trained neural network is treated as an oracle for the \HORN algorithm.
We implemented the algorithm in a Python~3.9 script and we used the SymPy library~\cite{10.7717/peerj-cs.103} to  
 express rules and check for satisfiability of formulas.
For the  neural networks, we used the Keras library~\cite{chollet2015keras}.
Our \HORN implementation can start with an empty hypothesis or with a set of Horn formulas as background knowledge
(assumed to be true properties of the domain at hand). 
%The algorithm will start with background knowledge that will help the learning process to ask less queries. 
The background knowledge can also be used to check 
if the neural network model respects some desirable properties. % that are wished to be respected by a learned model. %shortening this sentence 
%These can be any properties that can be expressed with Horn logic. 
We conduct the experiments on an Ubuntu 18.04.5 LTS with i9-7900X CPU at 3.30GHz with 32 logical cores, 32GB RAM.

We experiment our approach of extracting Horn theories from partial interpretations on a dataset in the medical domain~\cite{SANTOS201549}.
This dataset contains missing values for attributes. We can consider each instance as a partial
interpretation that sets some variables (attributes of that instance) to true, some to false, 
and other variables to ``unknown''. %or ``missing'' value. %, that correspond to the variables with missing values.

\subsection{HCC Dataset}
\label{ss:dataset}
Hepatocellular carcinoma (HCC) causes liver cancer, and it is a serious concern for global health. 
The HCC dataset~\cite{SANTOS201549}
consists of   $ 165 $ instances of many risk factors and features of real patients diagnosed with this illness. 
%In particular, it contains 

There are $ 49 $ features selected according to the EASL-EORTC (European Association for the Study of the Liver - European Organisation for Research and Treatment of Cancer).
From these features,
%The dataset contains 
$ 26 $ are quantitative variables, and $ 23 $ are qualitative variables. %, plus the binary class target value. 
Missing values represent $10.22\%$ of the whole dataset and only $8$ patients have complete information in all fields (4.85\%).
The target class of each patient is binary.
Each patient is classified positively if they survive after $ 1 $ year of having been diagnosed with HCC, and negatively otherwise.
$ 63 $ cases are labelled negatively (the patient dies) and $ 102 $ positively (the patient survives).
%The problem of class imbalance is solved when training the neural network from partial interpretations.
%
%Describe quantitative and qualitative variables
Quantitative variables describe, for example, the amount of oxygen saturation in the human body, the concentration of iron in the blood, or number of cigarettes packages consumed per year. The range of the values that each variable can assume varies, but it is specified. Qualitative variables can only have two different values in this dataset (either 0 or 1). Usually they describe categorical information such as if the patient comes from an endemic country, or if it is obese, etc. 

% how to split qualitative variables
The \HORNs algorithm expects to receive counterexamples in the form of a partial interpretation 
that specifies the truth values of   boolean variables. For this reason, we encode quantitative variables in a binary representation format.
The  interval of values of each quantitative variable is partitioned into three sub-intervals. 
These intervals divide the values of the quantitative variable into ``low'', ``middle'', and ``high'' values. 
For example, the interval of values of the variable that describes the number of cigarettes packages consumed by the patient per year
is $ [0,510] $ can be  partitioned   into $ [0,50],(50,200], (200,510] $.

%~ For each quantitative variable $ v_i $, we create three boolean variables $ v_i^1,v_i^2,v_i^3 $.
%~ In our experiments,
%~ $ v_i^k $  is true if the value of the quantitative variable $ v_i $ has a value   in the interval $ k $, otherwise it is false.
%~ To partition an interval  $ I=[a,b] $, we first compute the average value $ a(I) $ of $ I $, then we divide
%~ $ I $ into $ I_1^{a(I)} = \{ n\in I \mid n \leq a(I)  \} $ and $ I_2^{a(I)} = \{ n\in I \mid n > a(I)  \} $. Afterwards, we compute $ a(I_1^{a(I)}) $ and $ a(I_2^{a(I)}) $. Finally, the three partitions of 
%~ $ I $ will be $ [a,a(I_1^{a(I)})], (a(I_1^{a(I)}),a(I_2^{a(I)})], (a(I_2^{a(I)}),b] $.

%~ In this method, if the average value is in the middle of the interval then this 
%~ corresponds to dividing the interval into $3$. However, if the average is lower or higher than the middle 
%~ value then the described method will take this account.
%~ %The described method  for 
%~ Binarising quantitative variables works  fine in our experiments with this dataset but it has some drawbacks. 
%~ For example, when the values have high variance, there may not be any values in the dataset belonging to the  selected middle interval. 

The binarised dataset has in total $ 26*3 + 23 +1= 102 $ variables and it can be considered a set of partial interpretations. 
A missing value in the new dataset is denoted with `?' similarly as in the original one, otherwise the value is $ 1 $ ($ 0 $) if the variable is set to true (false). 
Each partial interpretation $ I $ matches a rule (not necessarily Horn) of the form
\begin{equation}
\label{eq:d1} 
(l_1 \land \cdots \land l_{n-1} )\rightarrow l_{n}  
\end{equation}
where each $ l_i $ is a positive literal if the variable $ i $ is set to true in $ I $ and false otherwise. 
The literal $ l_k $ is not present in the rule if $ l_k $ has a missing value.

As explained in the previous section, by duplicating the number of variables and pairing them such that one represents the negation of another variable, we can express the previous rule with a Horn formula.
For this reason, we further modify the dataset by duplicating variables. 
Each new variable semantically represents the negated concept of its paired variable.
%We assign to the new variables the opposite truth value of its paired variable. 
So, we form a dataset $ D $ of partial interpretations with $ 204 $ variables.

%Assuming the   disjoint constraints of the paired variables to hold (Formula~\ref{dualconstraint}), 
We can express 
each example in $ D $
with Horn rules like in Formula~\ref{eq:d1}.
%the previous non-Horn rule with a Horn rule
%that results from replacing
%each negative literal $\neg {\vsf}_{i}$ by  $ \hat{\vsf}_{i} $.
%\[ (\vsf_1 \land \cdots \land \hat{\vsf}_{n-1}) \rightarrow \vsf_{n}   \]
%where $ \hat{\vsf}_{i} $ denotes the new variable paired with $ \vsf_{n-1} $ that expresses its negation.
We denote by $ T $ the set of such rules that can be formed by looking at all partial interpretations in the extended dataset. 
To express disjointness constraints between paired variables,
we assume $ T $ to also have the additional Horn rules of the form
$( \vsf_i \land \hat{\vsf_i}) \rightarrow \bot $ (Formula~\ref{dualconstraint}).

Finally, the dataset used for training the neural network is formed by randomly generating partial interpretations (with $ 204 $ variables) whose classification label is $ 0 $ if they do not satisfy a rule in $ T $, $ 1 $ otherwise.

%\smallskip\noindent
%{\bfseries Model selection}
\subsection{Model selection}
 
By only randomly generating partial interpretations (with $ 204 $ variables), we can create a very
unbalanced dataset with most partial interpretations classified as positive 
by the target Horn theory   $ T $ (note: $T$ is defined in the previous subsection). 
We solve this problem by oversampling interpretations with negative label
that are created by violating rules that match interpretations in the binarised dataset.
In total, there are $ 200 $ negative examples and $ 200 $ positive examples in the training dataset.
$ 80\% $ of the (balanced) binarised dataset was used for training and validation. 
We used $ 3 $-fold.
As $ T $ is a Horn theory, there is no noisy data generated in this process.

We built a sequential neural network model, where the number of nodes in the
input layer is $ 204 $, which is the number of variables in a partial interpretation. 
We used the library
“Keras version 2.4.3” \cite{chollet2015keras} and we empirically 
searched for the sequential architecture with the best performance 
 varying the number of hidden layers, nodes in
hidden layers and the learning rate. 

%Table~\ref{parameters} shows the  taken into consideration together with 
We searched our model with the following hyper-parameters:
 2,3,4,5 numbers of hidden layers, 
 4, 8, 16, 32 nodes per layer, and
0.001, 0.01, 0.1 as the learning rate.
The model with the best performance 
had $ 5 $ hidden layers, $ 32,16,8,16,32 $ nodes per layer,
and $ 0.1 $ learning rate.
In total we tested 
(No.learning rates x No. node-layer combinations)
$= 3\cdot(4^2 +
4^3 + 4^4 + 4^5 ) = 4080 $  configurations.
This means that we carried in total $ 3 \cdot 4080  = 12240 $
training and evaluation runs. 
The best performing architectures are showed in Table~\ref{performing}.

%\begin{table}[t]
%%	\fontsize{9}{9}\selectfont
%	\begin{tabular}{ll}
%		\textbf{Hyper-parameter} & \textbf{Values} \\
%		No. Hidd. layers	& 2,3,4,5 \\
%		Nodes per layer & 4, 8, 16, 32 \\
%		Learning rate & 0.001, 0.01, 0.1 
%	\end{tabular}
%	\caption{The hyper-parameters used in the model selection process.}
%	\label{parameters}
%\end{table}

\begin{table}[t]
%	\fontsize{9}{9}\selectfont
	\begin{tabular}{lll}
		\textbf{Hidd. Layers} & \textbf{L. rate} & \textbf{Accuracy}  \\
		32, 16, 8, 16, 32   & 0.1 & 0.9612  \\
		32, 32, 32, 8   & 0.1 & 0.9592  \\
		32, 32, 32  & 0.1 & 0.9382  \\
		32, 8, 32, 16   & 0.1 & 0.9217  
		%		32, 8, 32, 16, 16  & 0.1 & 0.9783  \\
	\end{tabular}
	\caption{Architecture and learning rate of the top four neural networks in ascending order with respect accuracy. 
		The model   in the first row was the selected one.}
	%	 as it had the best accuracy. The accuracy has been rounded to four decimals. } % for use in the experiment.}
	\label{performing}
\end{table}

\begin{table}[t]
%	\fontsize{9}{9}\selectfont
	\begin{tabular}{lllll}
		\#Equiv. & t\_h & t\_nn & h\_nn &  t\_tree \\
		100           & 9.2\%        & 6.0\%   & 5.8\%   & 8.4\%        \\
	\end{tabular}
	\caption{The outcome of the rule extraction process with the HCC dataset. 
		The numbers are the percentages of interpretations classified differently between the target t (Section~\ref{ss:dataset}), neural network  nn, \HORNs hypothesis  h, and the tree.  }
	\label{hcc}
\end{table}

%\smallskip
%\noindent
%{\bfseries Test Setting.}
\subsection{Test Setting}
In our experiments, 
we run the \HORNs algorithm and we set a limit of $ 100 $ equivalence queries that the algorithm can ask before terminating with the built hypothesis as its output.
To simulate an equivalence query, we randomly generate a sample of partial interpretations and we  classify 
each interpretation using the neural network. Afterwards, we search for a counterexample to return to $ h $ as the answer of the query.

We compare the quality of the \HORNs hypothesis 
with the hypothesis formed by an incremental decision tree~\cite{DBLP:conf/kdd/DomingosH00}, an established white box machine learning model.
We use ``Hoefffding Decision Tree''  implementation 
present in the ``skmultiflow'' framework~\cite{skmultiflow}. 
It is possible to generate a set of propositional rules by visiting every branch of the tree from the root to leafs labelled negatively.
The sampling idea for finding negative counterexamples for \HORNs is also used for extracting a decision tree from the neural network.

We generate partial interpretations randomly and they are classified by the neural network.
We check if at least one of those classified partial interpretations is misclassified by the decision tree algorithm. If this is the case, we incrementally train the tree with the entire sample. This process is repeated until all classified interpretations in the sample are correctly classified by the tree.

Since the considered number of variables is $ 204 $,
it is not feasible to have the size of the sample for simulating equivalence queries
as dictated by 
Formula~\ref{sizesample} (this number is of the order $ 2^{204} $).
Moreover, a size of the sample too small often fails in finding a counterexample. 
When it is the case, the \HORNs algorithm will terminate and output a hypothesis with few (if not zero) rules,
and the tree will only be one node with label $ 1 $.
This problem is especially noticeable  in our current scenario as there are many variables 
but relatively few interpretations that are negatively labelled by the neural network.
The selected size of the sample is therefore
%\begin{equation*}
%\begin{array}{l}
%%$ 
%s := \left\lceil \frac{1}{\epsilon} log_2(\frac{{2^{|\Vsf|^{2.1} }}}{\delta}) \right\rceil,
%% $,
%\end{array}
%\end{equation*}
\[s := \left\lceil \frac{1}{\epsilon} log_2(\frac{{2^{|\Vsf|^{2.1} }}}{\delta}) \right\rceil\], 
both for training the decision tree and for answering queries asked by the \HORNs algorithm.

When the \HORNs hypothesis and the tree have been extracted, we 
compute a partial truth table of $ 204 $ variables of size $ 2s $.
We classify these interpretations according to the target $ T $,
the neural network, the \HORNs hypothesis and the decision tree.
We then compare the truth tables and count the number of times an interpretation is classified differently between the different models.
%We compute the number of interpretations
%labelled differently from a sample of random examples 
%of size $ 2s $. 

%\smallskip\noindent
%{\bfseries Results.}
\subsection{Results}
\Cref{hcc} shows the outcome of our experiment.
The columns
$ \emph{t\_h} $, $ \emph{t\_nn} $, $ \emph{h\_nn} $, $ \emph{t\_tree} $
are, respectively, the percentage of interpretations that are
labelled differently between the target and the hypothesis,
the target and the neural network, the hypothesis and
the neural network, and the tree and the target.
The running time of the \HORNs algorithm  with at most $ 100 $ equivalence queries was around $ 60 $ hours. The time for extracting an incremental decision tree is twice, around $ 120 $ hours.

The type of rules that the \HORNs algorithm extracted are of the form:
\[ \{ {\sf medium\_hemoglobin}  \land \cdots \land \hat{ {\sf obese}}  \rightarrow {\sf survives} \}\]
with around $ 40 $ different variables in the antecedent.
With $ 100 $ equivalence queries, the hypothesis extracted has 
$ 20 $ rules of this type that are also present in the target $ T $.
Other rules that are entailed by $ T $ can be found in the hypothesis.
Examples labelled negatively with many missing values contain more information about
the dependency between variables that must be respected.
Indeed, we noticed an increase of the accuracy of the neural network 
trained on more missing values ensuring
ensured balanced classes.
 As a consequence, 
also the quality of the extracted rules improves.

\section{Conclusion}
In this work we presented an approach for 
extracting Horn rules from neural network models
using partial interpretations.
It is often the case that not all values in a dataset 
are known or trustable. Our method based on partial interpretations
covers such scenarios and generalizes the case with (full) interpretations.
We test our approach empirically using a real world dataset in the medical domain.
%We performed experiments 

%We look forward to see you all in Troms{\o} for an engaging and productive conference!
\section*{Acknowledgements}
Ozaki is supported by the Research Council of Norway, project number 316022.

\bibliographystyle{abbrvnat}
\bibliography{references}

\end{document}